\documentclass[12pt]{article}

\usepackage[left=1in,top=1in,right=1in,bottom=1in,nohead,paperwidth=8.5in, paperheight=11in]{geometry} 
\usepackage{amsmath,amsfonts,amssymb,amsthm,dsfont }
\usepackage{geometry}
\usepackage{xcolor}
\usepackage{mathrsfs}
\usepackage{graphicx}
\usepackage{enumerate}
\usepackage{algpseudocode}
\usepackage{bbm}
\usepackage{hyperref}
\usepackage{float}
\usepackage{caption}
\usepackage[toc]{appendix}
\usepackage{algorithm}
\usepackage{setspace}
\usepackage{booktabs}
\usepackage{longtable}
\usepackage{natbib}
\usepackage[flushleft]{threeparttable}
\usepackage{tabularx}
\usepackage{enumitem}
\usepackage{graphicx}
\usepackage{subcaption}

\title{\bf Tree-based Forecasting of Day-ahead Solar Power Generation from Granular Meteorological Features
}

\author{Nick Berlanger$^{a,b}$\footnote{Corresponding author. E-mail: Nick.Berlanger@Luminus.be. 
The views expressed are purely those of the authors and can not, in any circumstance, be regarded as
stating an official position of Luminus.
Acknowledgments: IW was financially supported by the Dutch Research Council (NWO) under grant number VI.Vidi.211.032.}, Noah van Ophoven$^{c}$, Tim Verdonck$^{b}$ and Ines Wilms$^{c}$
		\\ \textit{\small $^{a}$ Optimization Department, Luminus, Brussels}
  	\\ \textit{\small $^{b}$ Department of Mathematics, University of Antwerp}
	\\ \textit{\small $^{c}$ Department of Quantitative Economics, Maastricht University}
}
\date{ }

\begin{document}

\def\spacingset#1{\renewcommand{\baselinestretch}%
	{#1}\small\normalsize} \spacingset{1}
	
\maketitle

\vspace{-12pt}
\noindent
{\bf  Abstract.}
Accurate forecasts for day-ahead photovoltaic (PV) 
 power generation are crucial  
to support a high PV penetration rate in the local electricity grid and 
to assure stability in the  grid. 
We use state-of-the-art tree-based machine learning methods to produce such forecasts and, unlike previous studies, we hereby  account for 
(i) the effects various meteorological as well as  astronomical features have on PV power production,
and this
(ii) at  coarse as well as granular spatial locations.
To this end, we use data from Belgium and forecast day-ahead PV power production at an hourly resolution.
The insights from our study can assist utilities, decision-makers, and other stakeholders in optimizing grid operations, economic dispatch, and in facilitating the integration of distributed PV power into the electricity grid. \\
\bigskip


\newpage
\spacingset{1.5} 

\section{Introduction} \label{sec:intro}
The integration of energy production from Renewable Energy Sources (RES) in the grid is a crucial pathway to the global reduction of greenhouse gas emissions and fossil fuel production \citep{ouik}.
In this paper, we focus on solar energy,  which is the second fastest-growing RES; indeed
the total installed photovoltaic (PV) power capacity in the world has increased from 42 GW in 2010 to 1 TW in 2022 \citep{ourworldindata_solarpv}.
However, despite the worldwide deployment of PV power and its contributions to a more sustainable future, the intermittent and volatile nature of solar energy poses considerably challenges to the operations of power grids.  
Accurate PV power forecasts then become crucial for energy suppliers to stabilize the grid, and to optimize solar unit commitments and economic dispatch \citep{dispatch}.
We consider tree-based machine learning methods to deliver such forecasts at the hourly resolution of the next business day, thereby accounting for meteorological and astronomical effects along a fine-grained  grid of spatial locations.

Electricity markets display several specific yet challenging attributes  compared to other commodities.
First, electricity  is economically non-storable as supply and demand need to be in constant balance. Additionally, demand and supply are intricately tied to external weather variables and calendar factors, introducing complex dynamics rarely seen in other markets \citep{balance}. 
With the rise of RES including PV energy, the complexity of the electric power systems is aggravating, as they become more dependent on the variability of external weather conditions.
As a consequence, electrical grids become more unstable, and the imbalance and instability between consumption and production increase \citep{jacobs}. 
Imbalances between power generation and demand can lead to increased costs for suppliers, including activating backup power plants or purchasing electricity at higher prices during periods of supply shortage \citep{lower_demand}. Conversely, when supply exceeds demand, suppliers can face disposal costs, including curtailing or selling excess electricity at lower prices, as well as the opportunity cost of unused generation capacity \citep{pv_maintainance_cost}. 
Accurate PV power forecasts are therefore  crucial for energy suppliers to balance their power portfolio and stabilize the grid. Besides, they also hold considerably value for developing clean energy power generation technology \citep{consumption}.

PV power forecast methods can be classified into two main categories: (i) physical methods and  (ii) data-driven methods (as well as combinations thereof), see \cite{solarreview2, recent_review}. 
Physical methods are built on analytical equations that characterize the PV power systems and typically use theoretical simulation models to calculate the output power of a PV system based on its main design parameters, irradiance forecasts and numerical weather predictions (NWP) (\citealp{duffie2020solar, masters2013renewable}).
Data-driven methods include  both  statistical, such as standard or penalized regression models \citep{linear, lasso} as well as machine learning (ML) methods that can capture complex non-linear relationships. ML methods for PV power generation forecasting range from state-of-the-art artificial neural networks (ANN, see e.g., \citealp{analysis_nwp_importance2, ann2, nn_short}), over support vector machines (SVM, see e.g., \citealp{svm_radial_best, svm_better_nn}) to tree-based methods (see e.g., \citealp{tree_models_outperform}). For a recent comprehensive overview, we refer the reader to \cite{recent_review}.
These approaches typically use historical PV data and external variables such as weather conditions 
to forecast future PV power output. 

Data-driven  methods have recently become more popular due to the availability of large data at solar stations.
Recent work by \cite{benchmark_1, ml_short_term_better}, and the review by \cite{ml_outperform_largedata} show that data-driven ML methods typically deliver the most accurate PV power forecasts at short-term  forecast horizons (up until 48 hours ahead). 
Among the ML methods, the review study in \cite{tree_models_outperform} as well as extensive empirical work by  \cite{benchmark_2, benchmark_1} point towards the superior performance of tree-based models, such as regression trees, random forest or boosting methods. These models also tend to be less complex and prone to overfitting compared to ANNs and SVMs \citep{tree_models_outperform}.

In this paper, we  therefore focus on tree-based ML methods in the important area of day-ahead PV power forecasting \citep{horizon}. Hourly day-ahead forecasts are mostly used for daily generation plans, economic dispatching plans, and trading on the day-ahead electricity market \citep{midlong}.
Our work contributes to this literature strand by offering a \textit{comprehensive forecasting framework}
that assists utilities such as energy suppliers and decision-makers in improving their grid operations, clean power energy technology, and lowering barriers for the integration of distributed PV in the local electricity grid.
Our framework is centered around three main pillars:
(i) we investigate  the performance of various \textit{tree-based ML methods},  ranging from regression trees, over random forest, to gradient boosting. We deliberately opt for tree-based methods (over neural networks or deep learning methods) not only due to their superior performance in the recent PV power forecasting literature but also thanks due their robust and sophisticated  implementations that are nowadays routinely available across standard software packages (see the recent discussion by \citealp{januschowski2022forecasting}), which greatly facilitates their adoption in industry. 
(ii) We use a distinct set of \textit{meteorological and astronomical features} combined with careful \textit{feature engineering} to forecast PW power. 
We contribute to the literature by providing an extensive discussion, based on a careful examination of the literature on solar engineering, photovoltaic systems and meteorology, of features that are deemed to be important in the forecasting model.
Finally,
(iii) we allow the effects of the various features to vary along a  \textit{granular spatial grid} of locations.  

We carefully assess forecast performance, from a statistical point of view by computing the Model Confidence Set \citep{mcs_hansen} to separate the best performing models from their competitors, as well as from a graphical point of view, by offering practitioners advice on a diverse set of  visualizations  one can use to assess different aspects of forecast performance. Furthermore,
to quantify the importance of meteorological and astronomical features at selected locations, we suggest to track feature importance based on SHAP (SHapley Additive exPlanation) values, as proposed by \cite{shap_founder}.

We apply our forecasting framework to Belgian hourly PV data over the period January 1st, 2019 to June 30th, 2023.
To our knowledge, no previous research has examined aggregated PV power output forecasts constructed at a regional (i.e.\ country) level and  from a diverse set of meteorological features available at a fine-grained spatial grid.
Besides, Belgium has not yet been the primary focus in earlier studies on PV power forecasting, yet its unique energy landscape makes it an especially pertinent subject for study. The country's ongoing energy transition, characterized by a move away from nuclear power and an increased emphasis on renewable sources, presents a compelling case for the necessity of accurate PV power forecasts. The Belgian climate, with its significant variability, underscores the importance of robust forecasting frameworks capable of handling such fluctuations. Furthermore, Belgium's availability of detailed hourly PV data makes it an ideal candidate for applying and evaluating the proposed forecasting framework; the results obtained can offer valuable insights for other regions facing similar integration challenges.

The remainder of this paper is organized as follows. 
Section \ref{section:data} introduces the data and describes the feature engineering steps. 
Section \ref{section:ML} discusses the ML methods that we use to forecast PV power. 
Section \ref{section:forecast_design} introduces our forecast framework, including the different model configurations, as well as the overall training, validation, and test set-up. 
Section \ref{section:result} presents the results, which comprise forecast performance of all models, and a feature importance study. 
Section \ref{section:conclusion} concludes.

\section{Data and Feature Engineering}\label{section:data}
We aim to obtain accurate day-ahead PV power forecasts. To this end, consider the following model 
\begin{equation}\label{equation 1}
    Y = f(X) + \varepsilon,
\end{equation}
where $Y$ is our target variable namely PV power, $X = (X_1, X_2, \dots, X_p)$ is a vector of $p$ features,  $f$ denotes the forecast function that connects the features to the target, and $\varepsilon$ is the error term.

Section \ref{subsec:PV_power} provides details on the PV power target, 
Section \ref{subsec:data_features} on the meteorological and astronomical features. 
Table \ref{table:data} presents an overview of all variables used throughout the paper, together with their abbreviations and  corresponding data providers. 
Our time frame spans from January 1st, 2019 to June 30th, 2023, resulting in a total of  $T = 39,407$ hourly observations. Note that all time units are measured in local time, thereby accounting for the seasonal shifts between summer and winter time in accordance with the Daylight Saving Time policy in Belgium.

\begin{table}[t]
    \centering
    \footnotesize
    \caption{Data description and data sources.}  \label{table:data}
    \begin{tabular}{p{2cm}p{6.5cm}p{6cm}}
    \toprule 
    Abbreviation & Variable Description  & Data Source
     \\
    \midrule
      ASG & Aggregated Actual Solar Generation Belgium (MW) & \cite{elia_solar_data}  \\ 
 IC & Installed Capacity in Belgium (MW) & \cite{elia_solar_data}\\
 SNR & Individual cell Surface Net Solar Radiation (J m$^{-2}$) &  \cite{cds} \\ 
 SSD &  Individual cell Surface Solar Radiation Downwards (J m$^{-2}$) & \cite{cds} \\
 T2m & Individual cell Temperature at 2m (Kelvin) & \cite{cds} \\
 RH & Individual cell Relative Humidity (\%) & \cite{cds} \\
 WCI & Individual cell Wind Chill Index ($^\circ$C)& \cite{cds} \\
 TCC & Individual cell Total Cloud Cover (0 - 1) & \cite{cds} \\
 Zenith & Zenith (Degree Angle) & (N/A)\\
 Azimuth & Azimuth (Degree Angle)& (N/A) \\
\bottomrule
\end{tabular}
\vspace{0.5px}
\begin{flushleft}
Note: The variables zenith and azimuth have no data source since they are constructed manually.
 \end{flushleft}   
\end{table}

\subsection{PV Power Output} \label{subsec:PV_power}
\begin{figure}[ht]
    \begin{subfigure}[b]{0.5\linewidth}
         \centering
         \includegraphics[width=0.85\linewidth]{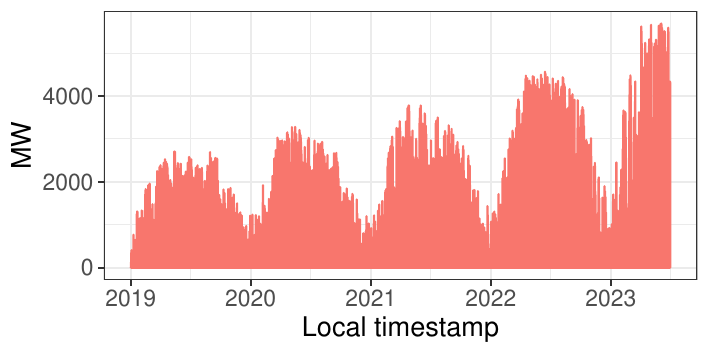} 
         \caption{ASG plot for the whole data set.} 
         \label{solar_power_ts_plot_01} 
         \vspace{4ex}
    \end{subfigure}
    \begin{subfigure}[b]{0.5\linewidth}
         \centering
         \includegraphics[width=0.85\linewidth]{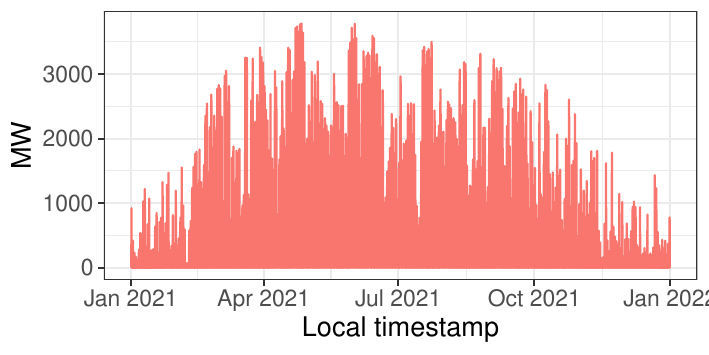} 
         \caption{ASG plot for 2021.}
         \label{solar_power_ts_plot_02} 
         \vspace{4ex}
    \end{subfigure} 
    \begin{subfigure}[b]{0.5\linewidth}
         \centering
         \includegraphics[width=0.85\linewidth]{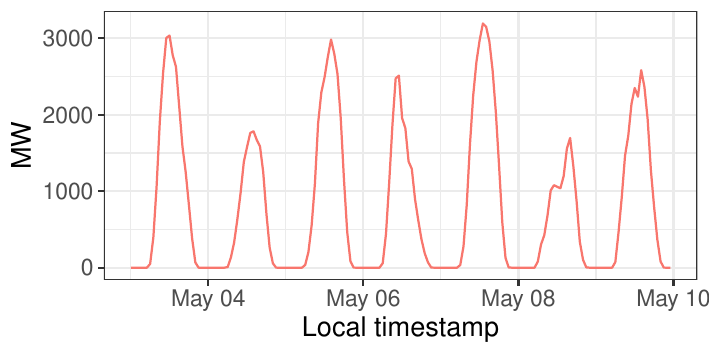} 
         \caption{ASG plot for May 3-9, 2021.} 
         \label{solar_power_ts_plot_03} 
    \end{subfigure}
    \begin{subfigure}[b]{0.5\linewidth}
         \centering
         \includegraphics[width=0.85\linewidth]{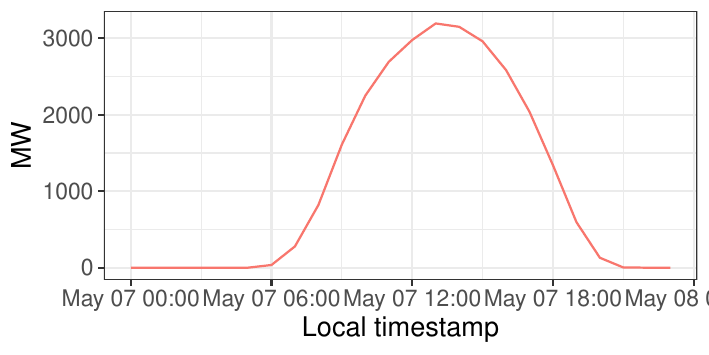} 
        \caption{ASG plot for May 7, 2021.} 
         \label{solar_power_ts_plot_04} 
    \end{subfigure} 
\caption{Actual Solar Generation (ASG) plots for different time periods.}
\vspace{5px}
\label{solar_power_ts_plots} 
\end{figure}

We collect aggregated Actual Solar Generation (ASG) in Belgium from \cite{elia_solar_data}, which is the Transmission System Operator (TSO) of Belgium. 
Figure \ref{solar_power_ts_plots} presents hourly ASG, measured in Megawatts (MW) units, over the whole sample as well as over selected sub-samples to highlight typical seasonal patterns.

From Figure \ref{solar_power_ts_plot_01}, one can see that there is an overall upward trend in Belgian ASG, due to the increase in PV power installed capacity over the last years \citep{secondfast}. 
Figure \ref{solar_power_ts_plot_02} highlights that  ASG is typically lower in  winter months than in summer months. Weekly  and daily patterns can be observed from respectively Figure \ref{solar_power_ts_plot_03} and \ref{solar_power_ts_plot_04}. Note, however, that these are also influenced by the time of the year.

\begin{figure}
 \centering
  \begin{minipage}{0.48\linewidth}
    \centering
    \includegraphics[width=1\textwidth]{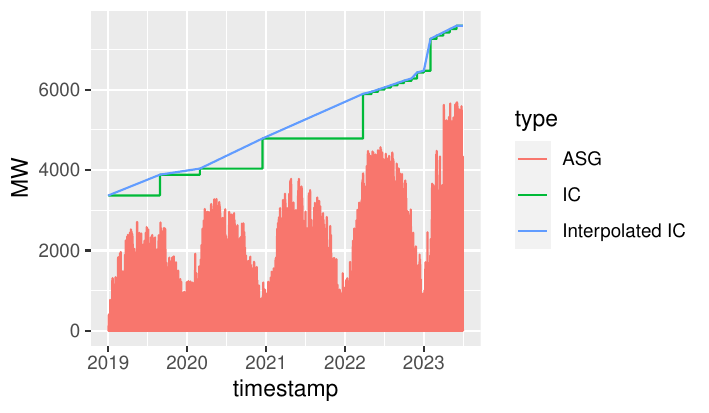}
    \caption{Construction of  the Load Factor.}
    \label{fig:target_engineering}
  \end{minipage}
  \hfill
  \begin{minipage}{0.48\linewidth}
    \centering
    \includegraphics[width=1\textwidth]{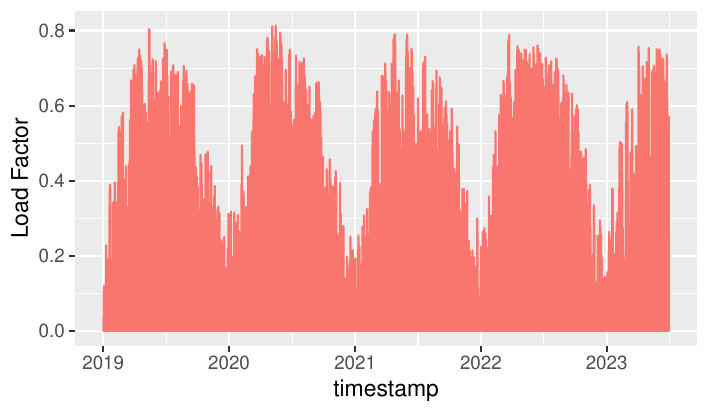}
    \caption{Load Factor (LF) time series.}
    \label{fig:load_factor}
  \end{minipage}
\end{figure}

We additionally collect Installed Capacity (IC) of PV power in Belgium in MW units from \cite{elia_solar_data}. Due to the granularity of the early data, we perform an interpolation, resulting in a smoothed capacity expansion curve illustrated in Figure \ref{fig:target_engineering} (with the original data represented by the green curve and the interpolated data by the blue curve). This approach ensures a more realistic depiction of capacity growth over time. The IC in Belgium increased considerably from 3,369 MW at the start of our sample to 7,593 MW at the end. 
For this reason, instead of forecasting raw ASG, we forecast a  normalized version of ASG, in line with \cite{bellinguer} who study wind energy output. 
By normalizing solar generation, the temporal evolution of the IC in Belgium is appropriately accounted for. 

The normalized ASG, also called Load Factor (LF), is then given by 
\begin{equation}\label{equation2}
    Y^{*}_t = \frac{Y_t}{IC_t} \;, 
\end{equation}
where $Y_t$ is the ASG at time $t$, $IC_t$ the interpolated IC at time $t$. Figure \ref{fig:target_engineering} shows all time series used to construct the load factor, which, in turn, is  visualized in  Figure \ref{fig:load_factor}. The resulting load factor time series is scaled between zero and one and no longer displays the increasing trend of the raw ASG.

\subsection{Features}\label{subsec:data_features}

\subsubsection{Meteorological Features}

\begin{figure}[t]
\centering
    \centering
    \includegraphics[width = 0.45\textwidth]{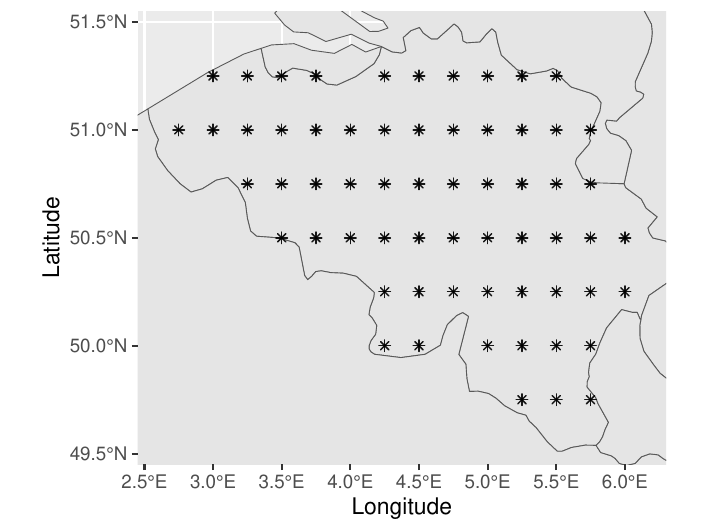}
     \caption{Grid points of the meteorological variables.}
     \label{fig:included_weather_locations}
 \end{figure}

\begin{figure}[H]
     \begin{subfigure}[b]{0.5\linewidth}
     \centering
     \includegraphics[width=0.9\linewidth]{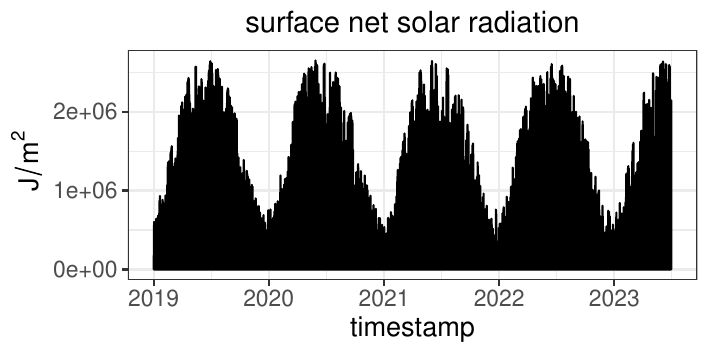} 

     \vspace{0.5ex}
   \end{subfigure}
   \begin{subfigure}[b]{0.5\linewidth}
     \centering
     \includegraphics[width=0.9\linewidth]{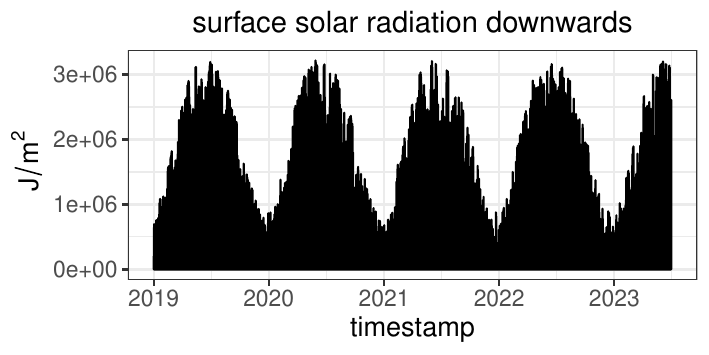} 

    \vspace{0.5ex}
   \end{subfigure} 
   \begin{subfigure}[b]{0.5\linewidth}
     \centering
     \includegraphics[width=0.9\linewidth]{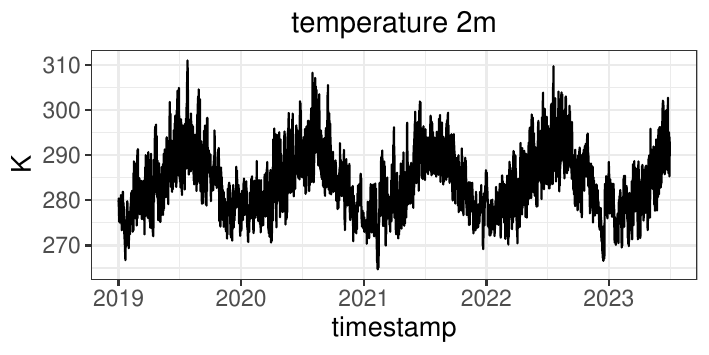} 
  
    \vspace{0.5ex}
   \end{subfigure}
   \begin{subfigure}[b]{0.5\linewidth}
     \centering
     \includegraphics[width=0.9\linewidth]{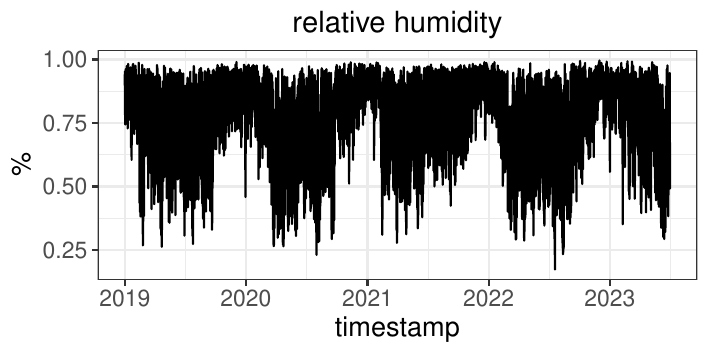} 
    
    \vspace{0.5ex}
   \end{subfigure} 
   \begin{subfigure}[b]{0.5\linewidth}
     \centering
     \includegraphics[width=0.9\linewidth]{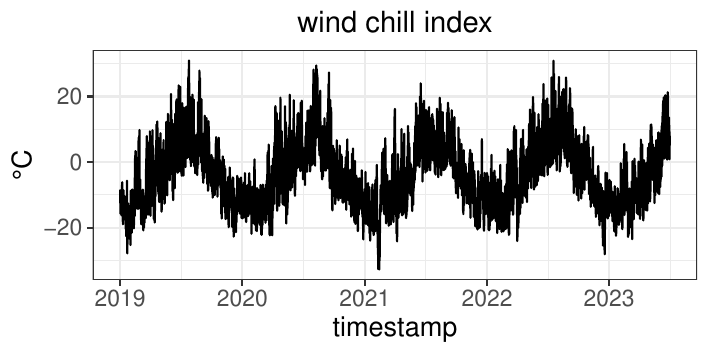} 
    \vspace{0.5ex}
   \end{subfigure}
   \begin{subfigure}[b]{0.5\linewidth}
     \centering
     \includegraphics[width=0.9\linewidth]{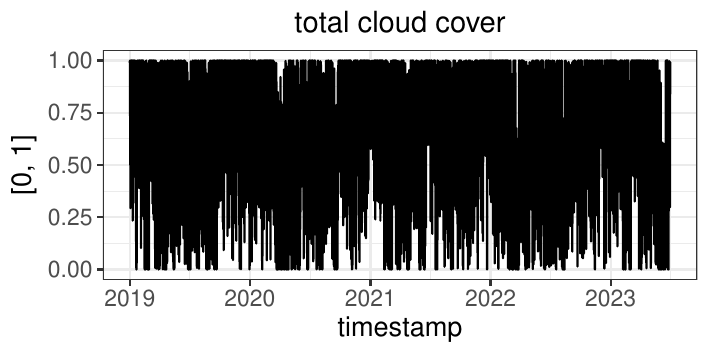} 
     \vspace{0.5ex}
   \end{subfigure} 
 \caption{Meteorological features averaged over the 62 locations depicted in Figure \ref{fig:included_weather_locations}.}
   \label{weather_time_series} 
 \end{figure}

We use six meteorological features
that are known to be important for forecasting PV power \citep{analysis_nwp_importance, linear, lasso}, namely 
Surface Net Solar Radiation (SNR), 
Surface Solar Radiation Downwards (SSD), 
Temperature at 2m (T2m), 
Relative Humidity (RH), 
Total Cloud Cover (TCC), and 
Wind Chill Index (WCI).
These features cover Belgium's spatial grid, including 62 locations as displayed in Figure \ref{fig:included_weather_locations}. 
Figure \ref{weather_time_series} presents the six meteorological time series, averaged over the 62 locations. 
Below we explain the specific relationship between each of these six meteorological features and PV power, thereby supporting their inclusion in the forecasting model. A detailed description of how these variables are measured can be found in \cite{Hersbach2023-cc, cds}.

\paragraph{Surface Solar Radiation.}
Surface Net Solar Radiation (SNR) represents the amount of solar radiation reaching the surface of the Earth minus the amount reflected by the Earth's surface, which is governed by the Albedo effect  \citep{albedo}. Indeed, radiation from the sun is partly reflected back to space by clouds and particles in the atmosphere, and some of the radiation is absorbed. Then, the difference between downward and reflected solar radiation is SNR.
Surface Solar Radiation Downwards (SSD),  on the contrary, gives the amount of solar radiation reaching the surface of the Earth and does not take into account the reflection of the sun by the Earth's surface. The units of SNR and SSD are Joules per square meter (J m$^{-2}$). We consider these features  since solar radiation has been shown to be the primary determinant of PV power generation \citep{analysis_nwp_importance, analysis_nwp_importance2}. Moreover, solar panels generate energy through the conversion of solar radiation, with direct solar radiation being regarded as one of the most promising sources of energy for this process \citep{solar_energy_how_generated}.

\paragraph{Temperature.} The variable T2m denotes the air temperature 2 meters above the Earth's surface, measured in Kelvin, which is a critical parameter for PV panel efficiency. \cite{non_data_temp} show that air temperature operates within a range that can optimize the output of a PV panel. While solar radiation increases air temperature by transferring heat energy, it is also essential to note the inverse relationship between PV efficiency and excessive heat. Elevated temperatures may reduce the operational efficiency of PV systems due to overheating of the panels \citep{temperature_efficiency}.

\paragraph{Total Cloud Cover and Relative Humidity.} The variable TCC quantifies cloud coverage as the fraction of the sky obscured by clouds within a grid box, while RH is defined as the ratio of actual vapor pressure to the saturation vapor pressure, expressed as a percentage.  
TCC is anticipated to adversely affect PV power generation since cloudier conditions lead to a reduction in solar irradiation compared to clear skies, thereby diminishing PV output \citep{non_data_tcc}. Similarly for relative  humidity, it is expected to indirectly impact PV output through its relation with solar radiation. Specifically, high RH can reduce the amount of sunlight that the panels can absorb, as photons are partially obstructed by denser water vapor in the atmosphere \citep{non_data_tcc}. Additionally, high humidity can lead to the formation of a thin layer of moisture on PV panels, further impairing their operational efficiency \citep{non_data_rh}. 

\paragraph{Wind Chill Index.} The last meteorological variable considered is WCI, which is a measure of the cooling effect of wind on the human body's perception of temperature. It is the perceived decrease in air temperature felt by the body on exposed skin due to the flow of air \citep{wind_chill_index}. While this measure is based on temperature and wind speed concerning human comfort, it has applicability in assessing environmental conditions for PV systems. Wind speed, a component of WCI, can influence PV power output positively. Higher wind speeds may lead to a reduction in PV cell temperature, acting as a cooling mechanism and potentially enhancing efficiency \citep{wind_effect}. Hence, by including WCI as an additional feature, the effects of wind speed  on PV power can be taken into account \citep{wci}.

\subsubsection{Astronomical Features}
Finally, we use two astronomical features namely Zenith and Azimuth  which are solar position variables utilized to track the position of the sun in the sky. Zenith represents the angle between the sun and the vertical direction of the observer, or alternatively, the angle between the sun and the point directly overhead. This angle is measured in degrees, where 0$^\circ$ corresponds to the point directly overhead and 90$^\circ$ represents the horizon. The Azimuth angle, on the other hand, is the angle between the due south and the projection of the sun's position onto the horizon of the individual. It is measured in degrees from 0$^\circ$ to 360$^\circ$ in a clockwise direction from the due south \citep{zenith_azimuth}.

The inclusion of Zenith and Azimuth as features in the PV output forecast model is important because they have an effect on the amount of radiation that reaches the surface of the earth. These variables explain the amount of radiation received at different times of the day and at different latitudes and thus directly impact the amount of solar energy produced at a given location \cite{solarpos_important}. Furthermore, these variables serve as a proxy for the temporal components or calendar effects that are related to ASG, as discussed in Section \ref{subsec:PV_power}. 

We compute the two astronomical variables in \texttt{R} \citep{r2023} using the \texttt{solarPos} package \citep{solarpos} which implements the solar position algorithm as described in \cite{zenith_azimuth}. The package calculates the position of the sun based on the latitude, longitude, and time of the observation in Julian time units. For our study, we apply this for the single location with coordinates $4.64$ $^\circ$E, $50.65$ $^\circ$N, as determined by averaging the longitude and latitude values from all locations depicted in Figure \ref{fig:included_weather_locations}. 

\section{Tree-based Machine Learning Methods}\label{section:ML}
This section reviews the  Machine Learning (ML) methods that we use to forecast PV power to make the paper comprehensive. 
Section \ref{subsec:RT} outlines regression trees, Section \ref{subsec:RF} random forest and Section \ref{subsec:XGBoost} (extreme) gradient boosting. 
For a more extensive introduction to  each of these methods, we refer the interested reader to \cite{esl} (Chapters 9, 10 and 15).

\subsection{Regression Trees} \label{subsec:RT}

Regression trees (RT) partition the features space into smaller and smaller regions, and use local averages in each region to forecast the response.    
Assume the space is split into $M$ regions $R_1, R_2,\dots,R_M$, the forecast function is then given by
    ${f}(X) = \sum_{m=1}^{M}c_m I\{X \in R_m\}\,,$
where the regression model predicts $Y$ with a constant $c_m$ in region $R_m$.
The estimate of $c_m$ is simply given by the average of the response values (in the training set) in region $R_m$, if one adopts the sum of squares as loss function.
The regression tree can then be used to forecast the response for new observations by traversing down the tree from the root to a leaf node, based on the values of the features, and taking the average of the response value in the corresponding leaf node as forecast.

To train the RT, we use  the greedy top-down recursive partitioning method explained in \cite{Breiman} and implemented in the \texttt{R}  package \texttt{rpart} \citep{rpart}. 
Regression trees have three important tuning parameters.
The first two, namely 
the  maximum tree depth ($\text{MaxDepth}$) and minimum number of observations per terminal (leave) node ($\text{min}_n$), are used as stopping criteria when building the tree.
The third one is the cost complexity parameter $\alpha$, which is used to prune the tree to reduce overfitting. 
 In Section \ref{subsection:forecast_design_splits}, we provide more details on how these parameters are tuned.

\subsection{Random Forest} \label{subsec:RF}
Random Forest (RF) is proposed by \cite{breimann_rf} as an ensemble method of multiple regression trees. 
Each tree is built by considering a random subset of features at each split of the tree, thereby providing diversity in the forecasts of the tree. 
By randomly selecting features as candidates for splitting, RFs can improve the variance reduction of bagging by reducing the correlation between the trees (thereby not increasing the variance too much).
After growing a specified number of trees, the final forecast is simply the average of the forecasts across the large collection of de-correlated trees.

We use the \texttt{ranger} package  \citep{rangerR} in \texttt{R} to train the RF. The tuning parameters are the number of trees, the number of features to randomly select from per split ($\text{mtry}$) and the minimum number of observations  per terminal node ($\text{min}_n$). We fix the number of trees at 600 and tune the other two parameters as will be detailed in  Section \ref{subsection:forecast_design_splits}.

\subsection{Extreme Gradient Boosting} \label{subsec:XGBoost}
Finally, we consider Extreme Gradient Boosting (XGBoost), as proposed by \cite{xgboost_paper}. XGBoost 
builds on the gradient boosting  approach of \cite{Friedman_AOS_2001} where  the forecast function is constructed through  a sequential ensemble of small regression trees, called weak learners, which together learn to build a strong learner. 
Each new learner is added based on the residual error obtained from the previous iteration of the weak learner. 
Gradient boosting then results in a one final tree, constructed as a weighted sum of trees, to be used for forecasting.
XGBoost  further optimizes the implementation of gradient boosting in terms of  flexibility and computing time.

We use the XGBoost algorithm available in the  \texttt{xgboost} package \cite{xgboost} in \texttt{R}. 
For fair comparison to RF, we set the number of trees to 600. 
The hyperparameters requiring tuning are  $\text{mtry}$, $\text{min}_n$, and $\text{MaxDepth}$ (as discussed in Sections \ref{subsec:RT} and \ref{subsec:RF}), the learning rate ($\eta$) which scales the contribution of each tree in the ensemble, the loss reduction ($\gamma$) which controls the model complexity of the trees, and finally SubSample which regulates the fraction of training data used for each boosting iteration to reduce overfitting. 

\section{Forecasting Framework}\label{section:forecast_design}

We present a comprehensive outline of our forecast framework. First, Section \ref{subsection:forecast_design_config} offers an overview of the various model configurations we consider. Second,  Section \ref{subsection:forecast_design_splits} outlines the design of the forecast study that is used for each model configuration. This includes the splitting of the sample into  training, validation, and test sets, the process of hyperparameter tuning on the validation set, and  the evaluation of forecast performance on the test set.

\subsection{Model Configurations}\label{subsection:forecast_design_config}
We consider 24 model configurations, where a model configuration refers to a combination of 
(i) ML method used to estimate the forecast function $f$ in equation \eqref{equation 1} (four considered methods), 
(ii) features used to predict the PV power output (two considered feature sets), and 
(iii) spatial grid of locations over which the  meteorological features are computed (three considered grids).  
For the remainder of this paper, we refer to a model configuration as a specific combination in these three dimensions.

\paragraph{ML Method.}
We investigate the performance of three different tree-based ML methods, namely regression trees (RT), random forest (RF) and extreme gradient boosting (XGBoost), see Section \ref{section:ML}, and compare them to the linear regression (LR) model which serves as a simple baseline.  All ML methods require tuning of their respective hyperparameters, which we discuss in Section \ref{subsection:forecast_design_splits}.

\paragraph{Feature Set.}
We consider two feature sets:
\begin{enumerate}[label=(\alph*)]
  \item Surface Net Solar Radiation (SNR), Zenith, and Azimuth; 
  \item Surface Net Solar Radiation (SNR), Surface Solar Radiation Downwards (SSD), Temperature at 2m (T2m), Relative Humidity (RH), Wind Chill Index (WCI), Total Cloud Cover (TTC), Zenith, and Azimuth.
\end{enumerate}
 Feature set (a) is the simplest specification and only includes three features that directly relate to solar energy via radiation. This specification is based on domain-expertise and supported by the main findings, summarized in Section \ref{subsec:data_features}, that solar radiation and solar position have the largest impact on the PV output. By adding Azimuth and Zenith angles, we include a proxy for the time patterns visible in ASG and LF. 
 In feature set (b), all meteorological variables introduced in Section \ref{subsec:data_features} are additionally included. 
 We are interested in investigating how these additional features  drive forecasting power.

\paragraph{Spatial Grid of Locations.}
We vary the number of spatial grid cells for all meteorological features to investigate the forecasting power of coarser as well as more granular grids. We consider three spatial grids.
In the first coarse case, for each meteorological variable, we take the simple average over all 62 locations shown in Figure \ref{fig:included_weather_locations}. 
For the other two more granular cases, we select specific grid points by $K$-means clustering, thereby varying the number of clusters $K$. The centroids $c$ from each of the resulting clusters $C$ indicate the most representative locations. We then select the grid points that are closest to these centroids as key locations. The  optimization problem is given by
\begin{equation}
    \underset{\;\;\;C}{\text{arg min}} \sum_{i=1}^{K}\sum_{x\in C_i} d_{h}(x, c_i)^2\;,
\end{equation}
where $x$ are the coordinates of the 62 grid cells, $c_i$ is the centroid of the points in cluster $C_i$, and $d_h(.,.)$ is the Haversine distance \citep{haversine}. The Haversine distance $d_h(.,.)$ between two coordinates $i$ and $j$ is calculated as
\[ 
\begin{aligned}
d_h(i,j) &= R \times 2 \times \text{{atan2}}\left(\sqrt{a}, \sqrt{1-a}\right),\\
\text{{where}}\;\, a &= \sin^2\left(\frac{\Delta\theta}{2}\right) + \cos(\theta_i) \times \cos(\theta_j) \times \sin^2\left(\frac{\Delta\lambda}{2}\right), \\
\end{aligned}
\]
with $\theta_i = \text{{lat}}_i \times \frac{\pi}{180}$,  $\theta_j = \text{{lat}}_j \times \frac{\pi}{180}$, $\Delta\theta = (\text{{lat}}_j - \text{{lat}}_i) \times \frac{\pi}{180}$, $\Delta\lambda = (\text{{lon}}_j - \text{{lon}}_i) \times \frac{\pi}{180}$, where $\text{{lat}}_i$ ($\text{{lon}}_i$) are the latitude (longitude) of coordinate $i$, similarly for coordinate $j$, and $R$ is the radius of the earth \citep{haversine}.

\begin{figure}[t]
 \centering
     \begin{minipage}[c]{0.46\linewidth}
     \centering
     \includegraphics[width = 1\textwidth]{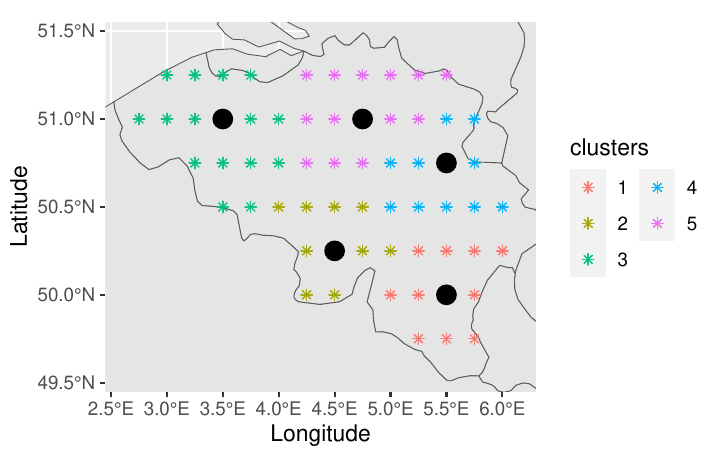}
     \caption{Grid points for $K=5$ clusters.}
     \label{fig:clusters_5}
     \end{minipage}\hfill
     \begin{minipage}[c]{0.46\linewidth}
     \centering
     \includegraphics[width = 1\textwidth]{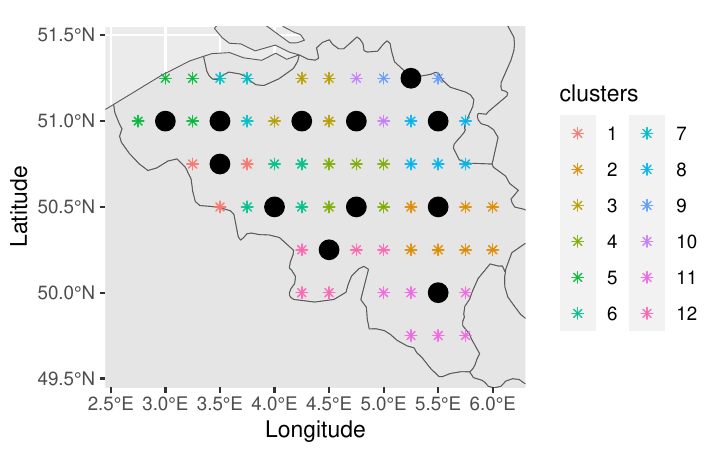}
     \caption{Grid points for $K=12$ clusters.}
     \label{fig:clusters_12}
     \end{minipage}
\end{figure}

For the granular spatial grids, the number of clusters is set to $K=5$ and $K=12$; the resulting clusters  with the corresponding key locations obtained are visualized in Figures \ref{fig:clusters_5} and \ref{fig:clusters_12} respectively. 
We choose these values of $K$ as both cover the whole territory of Belgium, while still considerably reducing the dimensionality of the problem. 
The $K=5$ locations case gives a rough representation and thereby represents a simpler setting where the main geographical regions in Belgium are covered (as will be discussed in Section \ref{subsection:result_shapley}), while the $K=12$ locations case aims to investigate whether the increased complexity in terms of spatial grid points pays off in terms of improved forecast accuracy.

\subsection{Design of the Forecast Study}\label{subsection:forecast_design_splits}
To conduct our forecast exercise and compare forecast accuracy across the 24 model configurations, we split the data into a training, validation, and test set. The training set is used for training the ML models, the validation set for hyperparameter tuning and the test set for out-of-sample forecast performance evaluation.

The (initial) training set ranges from January 5, 2019, to January 4, 2021 (731 days); 
the validation  set covers the period from January 5, 2021, to December 30, 2021 (360 days); and the  
test set encompasses the time period from January 1, 2022,
to June 30, 2023 (545 days).  

To optimize the tuning, training, and forecasting process, all night hours (when PV generation is typically zero) have been excluded from the dataset. Specifically, only the hours from 5 AM to 9 PM (hour 5 to 21) are retained. For the test set, which requires a complete 24-hour structure for evaluation, the omitted night hours are filled with zeros. This approach reflects the negligible solar power generation during these hours. Additionally, any negative forecasts generated by the models are adjusted to zero, as PV power output cannot be negative.

Finally, note that to validate and test the model, we intentionally included approximately one year and one and a half years of data, respectively, ensuring that the performance assessment and model selection include both winter and summer months. This approach allows us to account for the seasonal effects of ASG on forecast performance. 
We conduct our forecast exercise in \texttt{R}. The data set used in this paper as well as all replication files for the forecast study can be found on the GitHub page \url{https://github.com/nberl/tree-based-dah-solar-forecast} of the first author.

\begin{figure}
 \centering
     \begin{minipage}[c]{0.45\linewidth}
     \centering
     \includegraphics[width = 1\textwidth]{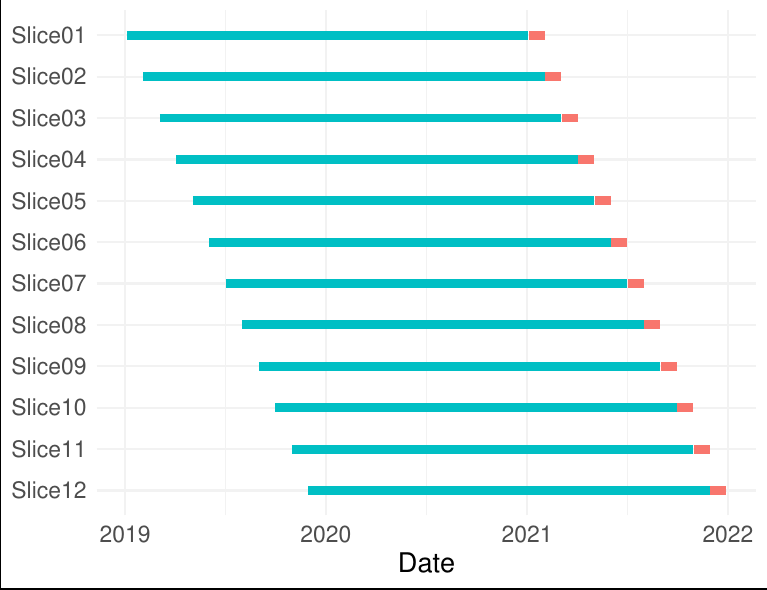}
     \caption{Rolling window procedure on the validation set.}
     \label{tune_validation_windows}
     \end{minipage}\hfill
     \begin{minipage}[c]{0.45\linewidth}
     \centering
     \includegraphics[width = 1\textwidth]{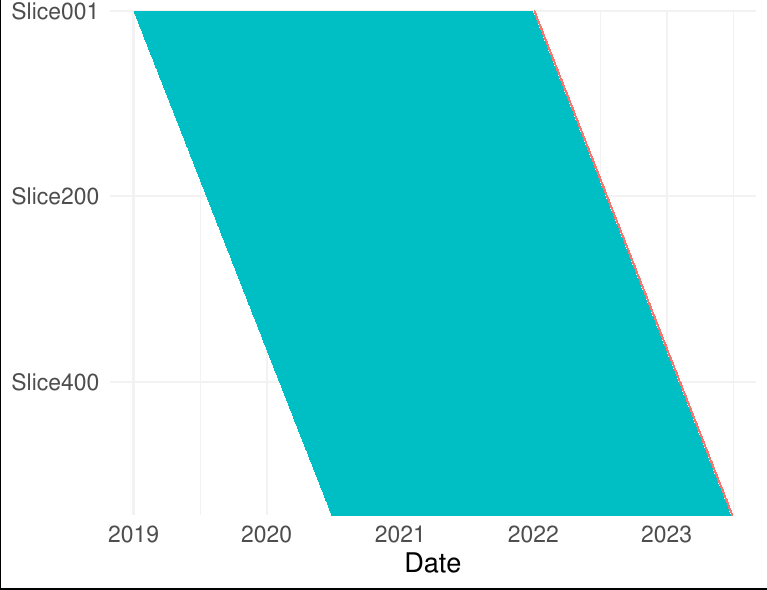}
     \caption{Rolling window procedure on the test set.}
     \label{train_forecast_windows}
     \end{minipage}
 \end{figure}

\paragraph{Validation Set: Hyperparameter Tuning.}
To tune the hyperparameters of the ML methods on the validation set, we use a rolling window approach.
We train the ML model on the most recent 731 days, and refresh the validation set every 30 days. The validation set thus consists of 12 consecutive slices, each spanning 30 days as visualized in Figure \ref{tune_validation_windows}. For each slice, the blue segment represents the training data whereas the red segment represents the validation data. 
When training a particular ML model, we do this using a grid of hyperparameters for the respective ML method-- RT, RF or XGBoost, see Section \ref{subsec:RT}. 
The hyperparameters of each ML method are summarized in Table \ref{list_hyperparameters}.
We start by selecting 25 candidate sets of hyperparameters by a space-filling Latin Hypercube Design (LHD). As such we  effectively explore the search space with a smaller number of evaluations compared to an exhaustive regular grid search \citep{latin_hypercube}. 
Specifically, a LHD partitions each hyperparameter grid into equally spaced intervals and then randomly selects a single value from each interval. The selected values form a LHD which ensures that the sampled points are evenly distributed across the hyperparameter space. The key idea is thus to obtain a representative sample of the hyperparameter space while reducing redundancy. 
We use the LHD implementation in the \texttt{dials} package \citep{latin_hypercube} in \texttt{R}. 
For each of the 25 candidates, we compute the root mean squared error (RMSE) across all time points of the validation sets, and finally select the best combination of hyperparameters which results in the lowest RMSE metric.  

\begin{table}[t]
    \centering
    \footnotesize
    \caption{Overview of hyperparameters per machine learning method.}
    \begin{tabular}{p{3cm}p{8cm}p{3.5cm}}
    \toprule 
    Abbreviation & Definition  &  ML Method(s)
     \\ \midrule
 $\alpha$ & Cost complexity for tree pruning & RT\\
 MaxDepth & Maximum tree depth &  RT, XGBoost\\ 
 $\text{min}_n$ &  Minimum observations per node for further split & RT, RF, XGBoost \\
 mtry & Number of randomly selected variables per split & RF, XGBoost \\
 $\eta$ & Learning rate or step size per ensemble iteration & XGBoost \\
 $\gamma$ & Penalty coefficient on number of leaf nodes & XGBoost \\
 SubSample & Fraction of training data used each iteration & XGBoost\\
 B & Number of base trees (fixed at 600) & RF, XGBoost \\
\bottomrule
\end{tabular}
\vspace{5px}
\begin{flushleft}
 \end{flushleft}   
    \label{list_hyperparameters}
\end{table}

\paragraph{Test Set: Forecast Evaluation.}
To compare the out-of-sample performance of the 24 model configurations on the test set, we use another rolling window set-up.
We train each ML model with selected hyperparameters on the most recent three years, and compute day-ahead 24-hour PV power forecasts. 
The first training set then ranges from January 1, 2019 (the start of our dataset) to December 30, 2021.
Afterwards in each iteration we move the training set one day forward until we reach the end of the sample. The first day in the test set is January 1, 2022.\footnote{ 
Note that there is always a gap of one day between the last observation date in the training data set and the forecast date since the date on which the model is trained and the forecasts are created are the same, but we do not have the complete data for that day available.}
The test set consists of 545 consecutive slices, each spanning 24 hours of the day as visualized in Figure \ref{train_forecast_windows}. For each slice, the blue segment represents the training data whereas the red segment represents the test data.

To evaluate the forecast performance of each model configuration, we compute three different forecast metrics across all time points of the test set. By comparing forecast performance across different metrics, we paint a broader picture when comparing the performance of the different model configurations. Specifically, we use the  root mean square error (RMSE), mean absolute error (MAE), and a symmetric mean absolute percentage error (SMAPE) respectively defined as
\begin{equation}\label{rmse}
    \text{RMSE} = \sqrt{\sum_{t=1}^{N}\frac{(\hat{Y}_t-Y_t)^2}{N}}\;, \ 
    \text{MAE} = \frac{\sum_{t=1}^N|\hat{Y}_t-Y_t|}{N}\;,\, \ \text{and} \ \ 
    \text{SMAPE} = 2 \frac{\sum_{t=1}^N|\hat{Y}_t-Y_t|}{\sum_{t=1}^N(\hat{Y}_t+Y_t)}\;, \nonumber
    \nonumber
\end{equation}
where $N$ denotes the number of time points in the test set, $\hat{Y}_t$ is the forecast of ASG at time $t$, and $Y_t$ is the  ASG value at time $t$. The three performance metrics have been used extensively in PV power forecasting \citep{metrics_pv_forecast, recent_review}. 
Note that each model forecasts the normalized ASG, as motivated in Section \ref{subsec:PV_power}, but the actual forecast quantity of interest is ASG.
We therefore  first transform the forecasts back to the raw ASG-level using equation (\ref{equation2}) before computing the forecast metrics.

\section{Results}\label{section:result}
We presents the results in two parts. 
Section \ref{subsection:result_out_of_sample} presents the out-of-sample forecast results of all 24 model configurations. 
Section \ref{subsection:result_shapley} zooms into the best performing model configurations and further compares them through a feature importance study based  on SHAP values \citep{shap_founder}.

\subsection{Forecast Performance}\label{subsection:result_out_of_sample}
We start by discussing overall forecast performance. 
Table \ref{overall_results} reports the out-of-sample RMSE, MAE, and SMAPE for the 24 tuned model configurations. 
A first glance at Table \ref{overall_results} reveals that, regardless of the chosen spatial grid and feature set, the ensemble methods RF and XGBoost obtain better forecast accuracy than the other methods, and this across all forecast metrics. 
Hence, non-parametric ML methods, where one does not have to worry about the functional form of the forecast function, and where feature selection and interaction are built-in, provide important forecast accuracy gains over simple, parametric LR models. This finding  is consistent with recent advances in the day-ahead forecasting literature.

\begin{table}[t]
    \centering
    \caption{Out-of-sample forecast performance for all 24 model configurations, defined through the combination of a particular ML method (rows), feature set (columns) and spatial grid (blocks across rows).}
    \begin{tabular}{ll ccc ccc}
    \toprule 
     & \multicolumn{4}{c}{\centering $\qquad \quad \qquad$Feature  Set (a)} & \multicolumn{3}{c}{Feature Set (b)} \\
    \cmidrule(lr){3-5} \cmidrule(lr){6-8}
    Spatial Grid & Method     & RMSE   & MAE  & SMAPE   & RMSE & MAE & SMAPE
     \\
    \midrule
    Average & LR       & 290 & 139 & 0.179  & 286  & 140  & 0.180   \\
     & RT  & 291 & 136 & 0.176  & 278  & 132  & 0.171   \\
     & RF   & 284 & 131 & 0.170*  & 266  & 131  & 0.169  \\
    & XGBoost  & 285 & 132  & 0.171  & 255  & 122   & 0.157  \\
    \midrule
   5 clusters & LR       & 285 & 137 & 0.177  & 277  & 139  & 0.178   \\
    & RT  & 290 & 137 & 0.178  & 284  & 135  & 0.175 \\
    & RF   & 275 & 128 & 0.166  & 252  & 119  & 0.154  \\
    & XGBoost  & 275   & 129 & 0.167  & 241*  & 117   & 0.151   \\
    \midrule
   12 clusters & LR       & 276 & 135 & 0.174  & 269  & 134 & 0.172   \\
    & RT  & 293 & 139 & 0.180  & 284  & 135  & 0.175 \\
    & RF   & 270 & 126 & 0.163  & 249  & 118  & 0.152**  \\
    & XGBoost  & 270  & 127 & 0.165  & 239**  & 115**  & 0.148  \\
    \bottomrule
    \end{tabular}
\vspace{5px}
\begin{flushleft}
  Note: The configurations included in the $99\%$ and $90\%$ MCS, computed per forecast metric separately, are identified by one and two asterisks respectively.
 \end{flushleft}   
    \label{overall_results}
\end{table}

When comparing performance across the feature  sets (a) and (b) and across the spatial grids,  
the differences in forecast performance are dependent on the ML method and the considered forecast metric. 
For the simple LR models, those models with feature set (b) perform slightly better based on the RMSE, but worse based on the MAE and SMAPE. Incorporating more locations in the linear regression model improves, overall, forecast performance. 
For the RT models,  in contrast, increasing the number of  spatial locations for meteorological variables generally leads to less accurate forecasts. For the best performing ML methods,  RF and XGBoost, they clearly show improved forecast performance on all forecast metrics by adding complexity in terms of features as well as spatial locations. 
When comparing the RF models to the XGBoost models, the former yields slightly more accurate forecasts for more simple models with feature type set (a), while XGBoost performs better for more complex scenarios, i.e.\ feature type set (b) and 5 or 12 clusters for the spatial locations.
Irrespective of the choice of forecast metric, the configuration XGBoost-(b)-12 yields the most accurate average forecasts on the considered test set, followed by the models XGBoost-(b)-5 and RF-(b)-12. 

We now globally assess the significance of the differences in forecast performance among the 24 model configurations by computing the Model Confidence Set \citep{mcs_hansen}. 
The MCS separates the set of best model configurations from their competitors for a given level of confidence.
We calculate the MCS for each forecast metric separately. For our analysis, each forecast metric is computed on an observation-by-observation basis. Consequently, we use the squared error instead of the RMSE and the absolute error for the MAE. In the case of SMAPE, where the denominator can be zero, we set the error to zero when the value is undefined. To calculate the results for the 99\% and 90\% MCS, we use the \texttt{MCS} package \citep{MCS_package} in \texttt{R}, thereby setting the  number of bootstrap iterations equal to 1000 and computing results for 99\% and 90\% confidence levels. The MCS is obtained for each forecast metric separately, model configurations included in the 99\% MCS ($\widehat{\mathcal{M}}^{\ast}_{99\%}$) and 90\% MCS ($\widehat{\mathcal{M}}^{\ast}_{90\%}$) are highlighted in Table \ref{overall_results} by respectively one and two asterisks. 

The model configurations  consistently excluded by all loss measures are configurations that do not employ either  XGBoost or RF as ML method. Under the RMSE loss, 2 models configurations are included in the $\widehat{\mathcal{M}}^{\ast}_{99\%}$, while only one configuration is included in the $\widehat{\mathcal{M}}^{\ast}_{90\%}$ set. 
The $\widehat{\mathcal{M}}^{\ast}_{99\%}$ set contains the XGBoost configurations with feature type set (b) with 5 and 12 clusters. 
The $\widehat{\mathcal{M}}^{\ast}_{90\%}$ set includes only the XGBoost-(b)-12 configuration, which displays the best performance across all model configurations. For the MAE loss, both $\widehat{\mathcal{M}}^{\ast}_{99\%}$ and $\widehat{\mathcal{M}}^{\ast}_{90\%}$ only include the XGBoost-(b)-12 configuration.
The SMAPE loss shows different results compared to the other loss metrics. $\widehat{\mathcal{M}}^{\ast}_{99\%}$ contains configurations RF-(a)-average and RF-(b)-12 and $\widehat{\mathcal{M}}^{\ast}_{90\%}$ only includes RF-(b)-12. This discrepancy arises due to the SMAPE being calculated on an observation-by-observation basis, combined with the nature of solar power data, which often features many near-zero values. Consequently, even small errors can lead to high SMAPE values when the denominator is small, questioning the appropriateness of SMAPE for hourly calculation in this context.

In summary, for  RF and XGBoost incorporating additional meteorological features such as T2m, RH, and TCC, in conjunction with additional spatial locations significantly enhance forecast performance. 
This is in contrast to the simpler ML methods that do not perform better in the more complex configurations; and are consistently excluded from the MCS. 
Next, we zoom into the forecasting performance of the top performing configurations.

\begin{figure}[t]
\centering
     \centering
     \includegraphics[width = 1\textwidth]{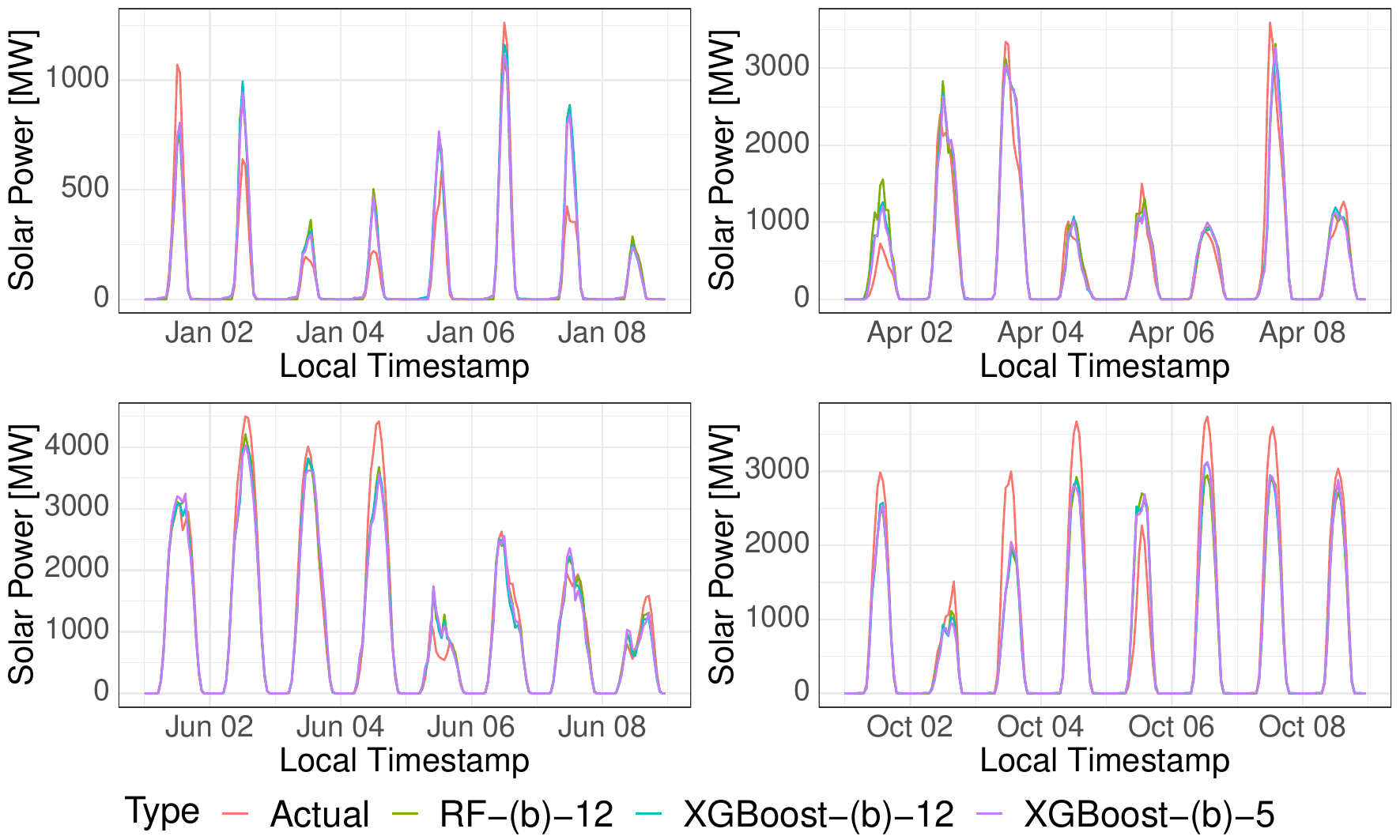}
     \caption{ASG forecasts for XGBoost-(b)-5, XGBoost-(b)-12, and RF-(b)-12 in 4 weeks during 2022 of the test set.}
     \label{example_forecasts_best_models}
 \end{figure}

Figure \ref{example_forecasts_best_models} displays the ASG forecast for the three best performing model configurations--XGBoost-(b)-12, XGBoost-(b)-5 and RF-(b)-12--across four representative weeks during 2022 of the test set.
These particular weeks are selected to give a more detailed insight into the day-ahead forecasts while covering different seasons throughout the test set.
The three model configurations provide very similar forecasts, especially for the beginning and end timestamps of the day, which also tend to be the most accurate time points. 
In contrast, the mid-day hour forecasts are less precise.

\begin{figure}[t]
 \centering
     \centering
     \includegraphics[width = 1.05\textwidth]{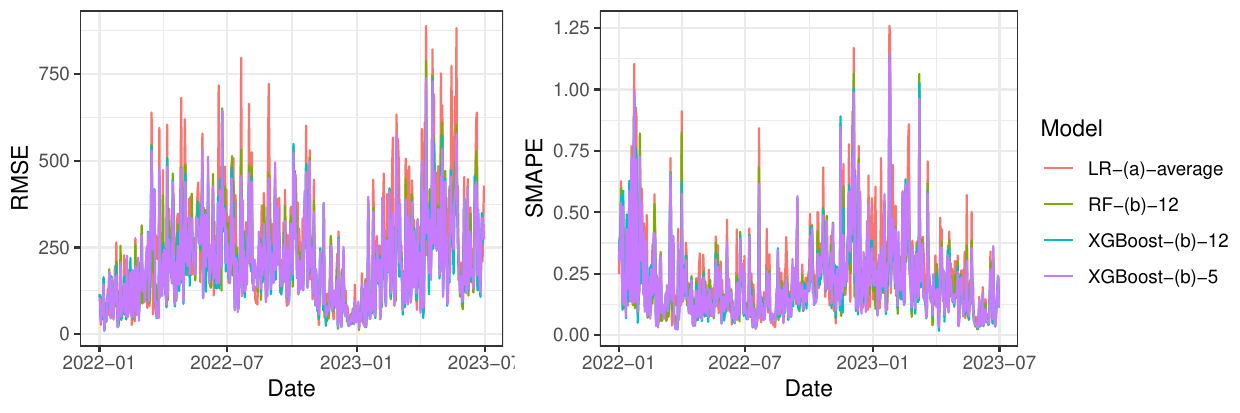}
     \caption{Daily RMSE (left) and daily SMAPE (right) plotted over the test-set for the configurations LR-(a)-average, XGBoost-(b)-5, XGBoost-(b)-12, and RF-(b)-12.}
     \label{daily_metrics}
 \end{figure}

\begin{figure}[H]
 \centering
     \centering
     \includegraphics[width = 1.05\textwidth]{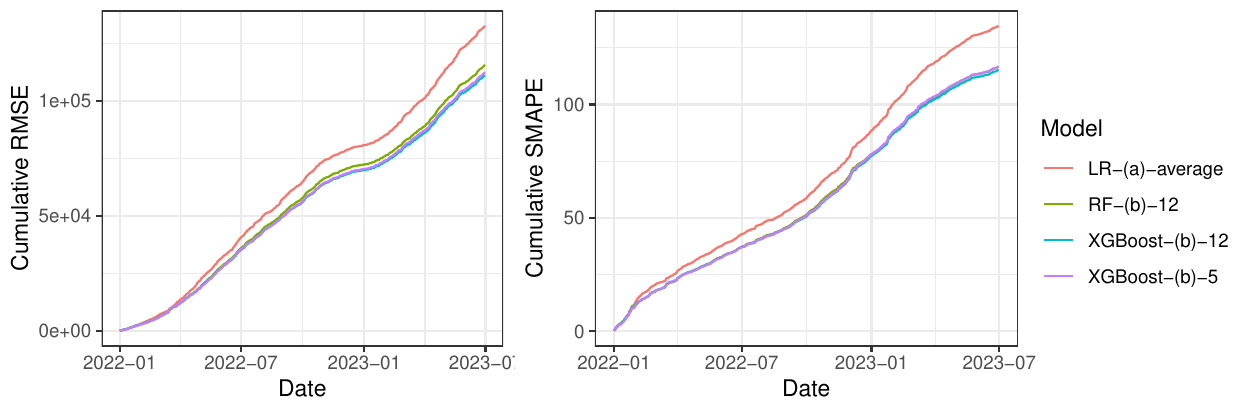}
     \caption{Cumulative daily RMSE (left) and daily SMAPE (right) plotted over the test-set for the configurations LR-(a)-average, XGBoost-(b)-5, XGBoost-(b)-12, and RF-(b)-12.}
     \label{cumulative_daily_metrics}
 \end{figure}

Figure \ref{daily_metrics} visualizes the daily RMSE and SMAPE performance of the top three configurations over the test set against the most simple configuration LR-(a)-average, whereas Figure \ref{cumulative_daily_metrics} displays their cumulative sums.\footnote{The findings for the MAE loss are similar to those of the RMSE loss and are therefore omitted.} 
From the RMSE and SMAPE plots in Figure \ref{daily_metrics}, we can clearly see that the linear model displays poor forecast performance with substantially higher peaks in the forecast errors particularly during the period from April to September. The cumulative plots in Figure \ref{cumulative_daily_metrics} reveal that the LR model initially performs similarly to the more complex model configurations until the end of May 2022 but thereafter starts to deteriorate considerably compared to the top configurations.
The top performing configurations all produce lower relative errors in the summer months when solar radiation is high and suffer from higher relative errors in the months from January to March and October to December when solar radiation decreases, see the SMAPE plot in Figure \ref{daily_metrics}. All three top configurations closely track each other throughout the whole test period as can be seen from Figure \ref{cumulative_daily_metrics} . 

The cumulative plots in Figure \ref{cumulative_daily_metrics} are also beneficial for detecting days or periods in which the model performs poorly compared to other periods. For instance, the performance of the top configurations during approximately the second half of March is very noisy with some inaccurate forecasts compared to the other forecasts in the test set. The slope of the cumulative plots during this period is indeed less smooth and some jumps can be observed. 
According to the national meteorological institution of Belgium responsible for monitoring and forecasting weather conditions, climate research, and providing meteorological services to the public and various sectors in Belgium \citep{KMI}, this result could be attributed to extremely cold and cloudy days at the end of March and the beginning of April, which deviated from the expected weather patterns for that time of the season. Additionally, as shown in Figure \ref{cumulative_daily_metrics}, a single jump can be seen on April 1st, 2022. \cite{KMI} provides insightful information regarding the atypical performance of the models in this day since  snowfall was recorded in multiple areas of Belgium, and this day was the coldest April 1st since the beginning of the measurements in 1901 thereby explaining why the forecast of ASG was considerably above the actual value. 
If one were to zoom into Figure \ref{cumulative_daily_metrics}, also July 21, 2022 and September 14, 2022, stand out as an atypical observations for a similar reason. Indeed,  \cite{KMI} reveals the former as the cloudiest day of July with the lowest amount of radiation and the second rainiest day of the month.
Similarly, September 14, 2022, was documented to be the cloudiest day of the month with the least solar radiation, 
thereby explaining the atypical performance of the models on this day. Given the exceptional nature of these three days, we classified them as outliers and excluded them from the tuning and training process to enhance model accuracy. However, they have been retained in the test set to evaluate the model's performance under atypical weather conditions.

A final important aspect to consider when comparing the top model configurations concerns computing time. Data science practitioners are nowadays increasingly challenged on their carbon footprints due to the running of computationally intense algorithms. Overall, the XGBoost models are less computationally complex than the RF model: for instance, training the best configuration XGBoost-(b)-12 and the configuration XGBoost-(b)-5 on the combined training and validation datasets takes approximately 2 minutes and 42 seconds, respectively, whereas the  RF-(b)-12 model configuration has a training time of 11 minutes.\footnote{Note that these training times are computed on a MacBook Pro (Retina, 15-inch, Mid 2014) with 2,2 GHz Quad-Core Intel Core i7 processor machine and only give an idea of relative training times.} These time differences further support the preference for choosing the XGBoost configurations over the RF-based model configuration as the former combine the best forecast accuracy with a substantial reduction in computational cost. 

\subsection{Feature Importance}\label{subsection:result_shapley}
In this section, we investigate which parts of the information set in the top model configurations generate predictability. To this end, we conduct a feature importance study based on SHAP values \citep{shap_founder}.  SHAP (SHapley Additive exPlanation) values are considered to be the current state-of-the-art method for interpreting ML models, as they offer a unified framework for comparing feature importance across different ML models. 
SHAP values are  based  on the principles of cooperative game theory \citep{Shapley_original} where the Shapley value is a unique solution under certain properties to a coalitional game in which the goal is to distribute the worth of a grand coalition among players in a fair way.  In the context of explaining ML model forecasts, the forecasts form the pay-off and the predictors are the individual players. The SHAP value for the $k$th feature at time point $t$ represents the feature's contribution to the forecast at that time point, as measured in terms of the deviation from the response mean over the training sample. A variable importance score for each feature can then be obtained by summing the absolute SHAP values over all desired time points. 
We refer the interested reader to \cite{molnar2020interpretable} for a detailed introduction to SHAP values.

To compute SHAP values for the RF and XGBoost models, we use the  \texttt{vip} package \citep{vipR}, which offers a fast implementation for tree-based ML models \citep{lundberg2020local}. Since each model is retrained every day, feature importance can change from day to day. To get an overall view of the SHAP values over the test set period, we calculate the SHAP values at the end of each month based on the forecast of the next month. Since our test set contains 18 months, the SHAP values are calculated 18 times. For each individual feature, we then compute its overall importance by taking the mean over all months of the test set. We subsequently analyze variable importance for each
(i) location in the spatial grid, by summing feature importance over all meteorological variables at the particular location, 
(ii) meteorological feature, by computing the mean over all considered locations, and finally each
(iii)  feature-location combination at the most granular level. 

\begin{figure}[t]
     \begin{subfigure}[b]{0.5\linewidth}
     \centering
     \includegraphics[width=0.9\linewidth]{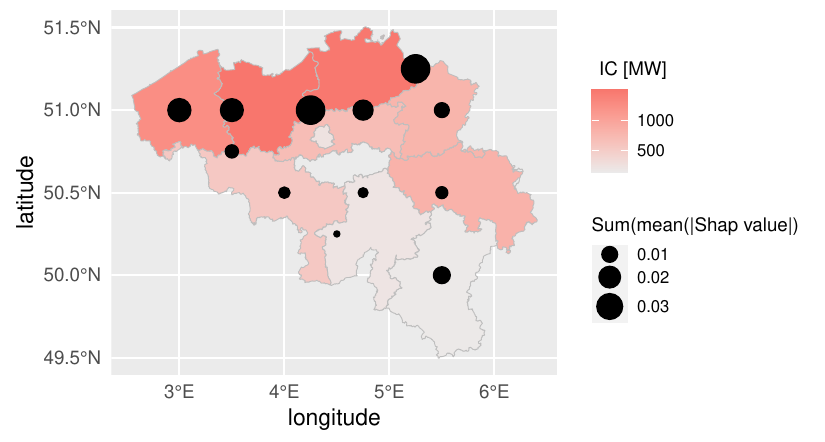} 
     \caption{ XGBoost-(b)-12 clusters} 
     \label{location_importance_xgb_b_12} 
   \end{subfigure}
   \begin{subfigure}[b]{0.5\linewidth}
     \centering
     \includegraphics[width=0.9\linewidth]{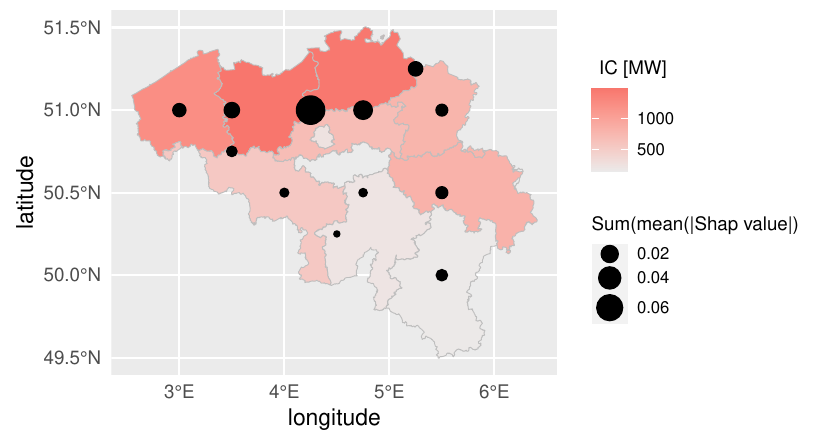} 
     \caption{ RF-(b)-12 clusters} 
     \label{location_importance_rf_b_12}   
   \end{subfigure}
   \caption{Location importance for the configurations with 12 clusters and feature  set (b).}
\label{location_importance_shap} 
\end{figure}

\paragraph{Location Importance.}
 Figure \ref{location_importance_shap} visualizes the contribution of each location to the overall forecasts for the two top performing model configurations RF and XGBoost with feature set (b) and 12 locations.
 Additionally, the Installed Capacity (IC) data of the provinces in Belgium, obtained from \cite{elia}, is indicated by the red shading in Figure \ref{location_importance_shap}. 
Figure \ref{location_importance_xgb_b_12} for XGBoost shows that the locations around the latitude of $51^\circ$N are most important. These locations correspond to the northern provinces of Belgium where the IC is substantially higher than in the southern provinces.
The most important locations at coordinates $4.25^\circ$E, $51^\circ$N  and $5.25^\circ$E, $51.25^\circ$N are roughly located in the two regions with the highest IC. 
These findings therefore add to the  validity of the feature importance study.

Figures \ref{location_importance_rf_b_12} visualize the spatial importance for the RF model. Overall, a similar conclusion can be drawn as for the XGBoost model, but the discrepancies between the locations are larger for the RF model. Indeed, while the location at $4.25^\circ$E, $51^\circ$N is the most important one in both the XGBoost and RF models, the XGBoost model distributes considerably more weight among other locations, particularly western grid points at the latitude of $51^\circ$N, compared to the RF model.

\begin{figure}[t]
     \begin{subfigure}[b]{0.5\linewidth}
     \centering
     \includegraphics[width=0.9\linewidth]{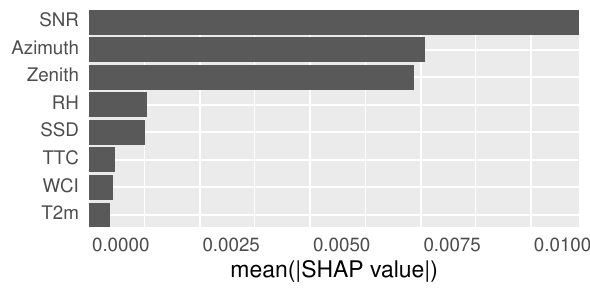} 
     \caption{ XGBoost-(b)-12 clusters} 
     \label{feature_type_importance_xgb_b_12} 
   \end{subfigure}
   \begin{subfigure}[b]{0.5\linewidth}
     \centering
     \includegraphics[width=0.9\linewidth]{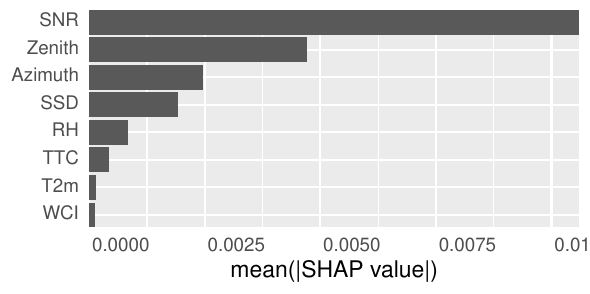} 
     \caption{ RF-(b)-12 clusters} 
     \label{feature_type_importance_rf_b_12} 
   \end{subfigure}
   \caption{Feature importance for the configurations with 12 clusters and feature  set (b).}
   \label{feature_type_importance} 
\end{figure}

\paragraph{Meteorological and Astronomical Feature Importance.}
Figure \ref{feature_type_importance} summarizes the meteorological and astronomical feature importance of the RF and XGBoost models with feature set (b) across 12 locations.
The SHAP results for XGBoost indicate that SNR is the most important for forecasting ASG, followed by Azimuth, Zenith, and RH, respectively. RF shows a slightly different feature importance where SNR is followed by Zenith, Azimuth and SSD. Hence, in line with the solar engineering literature \citep{duffie2020solar, masters2013renewable}, features directly related to solar energy have largest impact on day-ahead ASG forecasts. It is noticeable that Zenith and Azimuth display a similar feature importance in the XGBoost model configuration, whereas their importance differs substantially in the RF model configuration. The least important features are TCC, WCI, and T2m.

\begin{figure}[ht]
     \begin{subfigure}[b]{0.5\linewidth}
     \centering
     \includegraphics[width=0.955\linewidth]{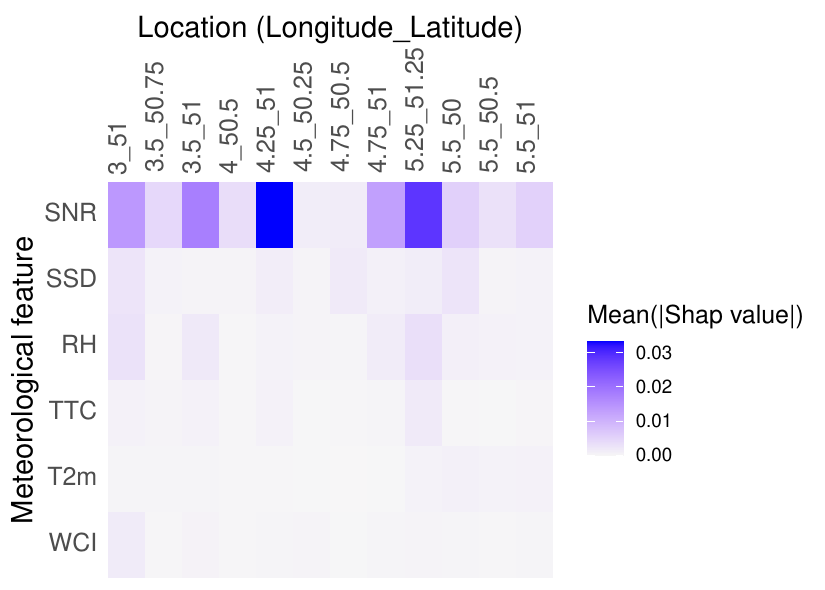} 
     \caption{ XGBoost-(b)-12 clusters} 
     \label{variable_importance_heatmap_xgb_b_12} 
   \end{subfigure}
  \begin{subfigure}[b]{0.5\linewidth}
     \centering
     \includegraphics[width=0.955\linewidth]{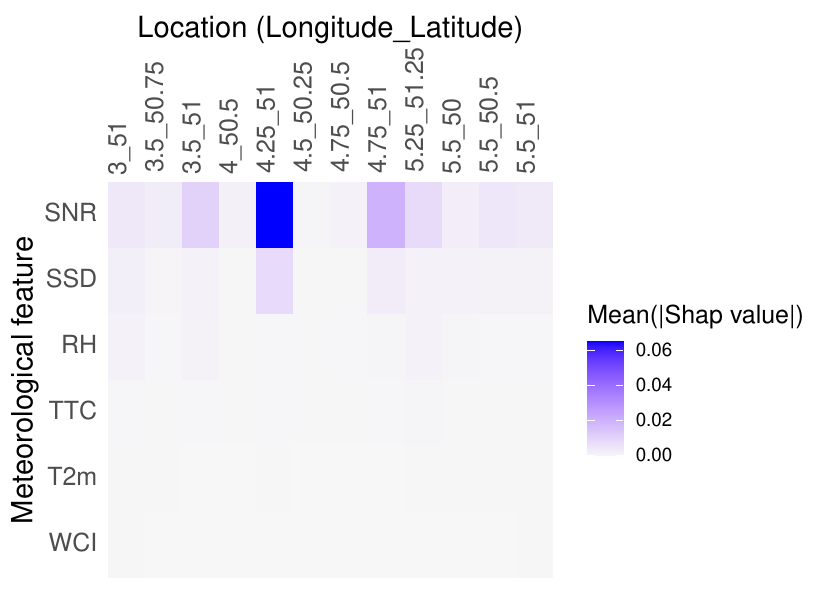} 
     \caption{ RF-(b)-12 clusters} 
     \label{variable_importance_heatmap_rf_b_12} 
   \end{subfigure}
  \caption{Specific feature importance for configurations with 12 clusters and feature set (b).}
   \label{variable_importance_heatmap} 
 \end{figure}

\paragraph{Feature-Location Specific Importance.}
Figure \ref{variable_importance_heatmap} illustrates the specific feature contributions at each spatial location. Note that Zenith and Azimuth are not included in this spatial analysis, as they are measured at a single location in Belgium. 
The main insights from the location and feature importance study are also reflected in the heatmaps of Figure \ref{variable_importance_heatmap}. The location at coordinate $4.25^\circ$E, $51^\circ$N is by far the most important one, especially with respect to  SNR and SSD. The heatmaps further reveal that XGBoost  distributes the SHAP values more evenly among all features than the RF model does, thereby  indicating that XGBoost utilizes a broader range of features to explain ASG. 

In summary, our feature importance study clearly reveals the need to include 
location-specific meteorological information in the ML model to achieve accurate forecast performance. 
A fine-grained spatial feature importance study hereby helps to identify key locations, which, in turn, may be useful to to evaluate the potential importance of specific sites \citep{site_importance}.

\section{Conclusion}\label{section:conclusion}
Due to the increased competition on the energy market and rise
of technological advances such as digital meters, it becomes  important for electricity providers and generators to correctly manage financial risks through models that capture electricity market
dynamics at a \textit{granular} level, as it would allow them to move towards a risk-based
approach of pricing electricity contracts.
This paper hereby focuses on forecasting short-term solar photovoltaic (PV) power, the second-fastest growing renewable energy source due to the increased penetration of PV energy into the electricity grid. 

We put forward a comprehensive  framework for forecasting day-ahead PV power generation using tree-based machine learning (ML) methods. The framework hereby accounts for possibly non-linear effects and the differential predictive power that various meteorological and astronomical features at fine-grained spatial locations have for (country-) aggregated Actual Solar Generation (ASG).
Employing such a granular meteorology-informed PV power forecast model is crucial for grid management, economic dispatch optimization, clean energy technology, day-ahead electricity market trading, and facilitating the integration of distributed PV power into local electricity grids.

Our application for Belgian ASG reveals that an ensemble-based ML method like XGBoost is key to obtain accurate forecasts. The XGBoost model performs best when a diverse set of meteorological and astronomical features-- whose relevance is supported by the solar engineering, photovoltaic systems and meteorology literature --measured at a fine-grained spatial resolution are included in the forecaster's information set. 
To gain insights into the forecast capabilities of various model configuration, we offer practitioners a diverse set of visualizations one can use to track forecast performance across time, and to reveal feature importance through SHAP values and this at a coarse  (location or feature-based) as well as at a fine-grained level (for location-specific features).

Several directions for future research emerge. We focus on solar-based energy, but the proposed forecasting framework could equally well be applied to model day-ahead wind energy output. 
Furthermore, we focus on point forecasts but probabilistic forecasts  could provide crucial additional information enabling decision-makers to assess risks and investments associated with resource allocation, grid stability, day-ahead trading, and other fields of PV power forecasting  \citep{probabilistic}. To this end, the quantile regression forest  approach of \cite{bellinguer} or the XGBoost extension by \cite{xgboost_quantile} offering forecast intervals and quantiles are interesting avenues to further explore.

Finally, while we focus on a short-term (day-ahead) forecast horizon, the ML models all  use  fundamental driving factors of solar generation,  selected based on a careful investigation of the meteorological literature, in the forecaster's information set. This opens up the door for future research into a holistic modeling approach between short-term and long-term models for electricity prices which crucially rely on expectations of renewable production and are essential for electricity providers to assess their risk exposure. 
To this end, it will be vital to investigate in future research how one can incorporate uncertainty regarding these fundamental drivers through scenario building  (see e.g., \citealp{muller2015modeling} for gas), and to robustify the forecasting framework to downweight the influence of atypical data values  in order to capture stable/robust long-term dynamics \citep{leoni2018multivariate}.

\begingroup
\bibliographystyle{apalike}
\setstretch{0.05}
\linespread{0.5}
\bibliography{bibliography}
\endgroup
\end{document}